\documentclass[runningheads]{llncs}
\usepackage{graphicx}
\usepackage{amsmath,amssymb} 
\usepackage{color}
\usepackage{comment}
\usepackage{booktabs}
\usepackage{multirow}
\usepackage{subcaption}
\usepackage{here}
\begin{document}
\pagestyle{headings}
\mainmatter

\def\ACCV22SubNumber{296} 

\title{Few-shot Adaptive Object Detection\\
with Cross-Domain CutMix}

\titlerunning{Few-shot Adaptive Object Detection with Cross-Domain CutMix}
\authorrunning{Y. Nakamura et al.}

\author{Yuzuru Nakamura\inst{1}\thanks{equal contribution} \and
Yasunori Ishii\inst{1\star} \and
Yuki Maruyama\inst{1} \and \\
Takayoshi Yamashita\inst{2}}

\institute{Panasonic Holdings Corporation
\email{\{nakamura.yuzuru,ishii.yasunori,maruyama.yuuki\}@jp.panasonic.com}\\
\and
Chubu University
\email{takayoshi@isc.chubu.ac.jp}
}


\maketitle
\begin{abstract}
In object detection, data amount and cost are a trade-off, and collecting a large amount of data in a specific domain is labor-intensive.
Therefore, existing large-scale datasets are used for pre-training.
However, conventional transfer learning and domain adaptation cannot bridge the domain gap when the target domain differs significantly from the source domain.
We propose a data synthesis method that can solve the large domain gap problem.
In this method, a part of the target image is pasted onto the source image, and the position of the pasted region is aligned by utilizing the information of the object bounding box.
In addition, we introduce adversarial learning to discriminate whether the original or the pasted regions.
The proposed method trains on a large number of source images and a few target domain images.
The proposed method achieves higher accuracy than conventional methods in a very different domain problem setting, where RGB images are the source domain, and thermal infrared images are the target domain.
Similarly, the proposed method achieves higher accuracy in the cases of simulation images to real images.
\end{abstract}

%
%
\section{Introduction}
Systems used in various environments throughout the day, such as autonomous driving and surveillance robots, are being put to practical use.
These systems require high accuracy throughout the day.
Infrared cameras can capture visible images even in such situations, and they can robustly detect objects.
To achieve high accuracy, object detection models require a large amount of labeled training data.
It is easy to use a large amount of labeled training data of RGB images~\cite{lin2014microsoft,Everingham15,openimages,shao2019objects365,kim2018textual,Cordts2016Cityscapes}.
However, most of the images are captured using an RGB camera, and a few images are captured using an infrared camera (hereinafter infrared images).
Therefore, to achieve high accuracy, detection models need to train with a few labeled infrared images.

One of the methods for training high accuracy detection model is transfer learning using pre-trained model trained with RGB images.
However, it is difficult to improve the accuracy if the domain gap between RGB and infrared images is large~\cite{pan2009survey,weiss2016survey}.
This phenomenon, called negative transfer, occurred under the large domain gap between the training images for pre-trained model and the ones for fine-tuning model.
As one of the conventional methods for overcoming the domain gap,
there are style transformation methods such as CycleGAN~\cite{zhu2017unpaired,shang2021survey}.
GAN-based style transformations can easily convert between images with similar spectra.
However, the style transformation is difficult when the spectral distributions are significantly different, such as in RGB and infrared images.
A method using GRL~\cite{osumi2019domain,bolte2019unsupervised,ganin2015unsupervised}
 is proposed as a training method for domain adaptation in the feature space.
However, if the spectral distributions between the input images are significantly different, it is difficult to align the distributions of different domains because the feature distributions are extremely different.
The methods for training features that interpolate between two images are as follows: Mixup~\cite{zhang2018mixup}, BC-learning~\cite{tokozume2018between}, CutMix~\cite{yun2019cutmix}, and CutDepth~\cite{ishii2021cutdepth}.
These methods train features located between two images by mixing the two images or by replacing a portion of the image with the other image.
These data augmentation methods synthesize features with a mixture of different domains.

We propose a few shot adaptive object detection with cross-domain CutMix.
We take advantage of the fact that data augmentation can reduce the domain gap by mixing features of domains with a large domain gap.
Our method enables highly accurate object detection even for a few annotated infrared images based on a pre-trained model of RGB images.
We paste a part of one domain's image onto a part of another domain's image, such as CutMix, because we overcome large domain gap.
Particularly, in object detection task, the size of detection targets is smaller than that of the background.
Therefore, to perform domain adaptation of small detection object, we cut out the detection object and paste it onto the other domain instead of randomly cutting out the image.
Even if there is a significant difference in appearance between domains, the features of the detected objects between domains are trained to be similar to each other.

Additionally, we adapt the domain using feature-based adversarial learning.
In conventional methods, the discriminator of the domain identification label does not change during training.
However, the domain identification label also needs to change the label according to the pasted area because our CutMix-based method changes the pasting area during training.
The conventional domain identification label cannot be used by pasting an image of another domain.
Therefore, the domain identification label should be the same domain label as the input image to which the image of another domain is pasted.
Since the correct domain label can be assigned according to the pasting position of the object, feature-level domain alignment can be performed even when the input image is changed, as in the proposed method.

Our contributions are as follows:
we propose a few shot adaptive object detection with cross-domain CutMix so that we can adapt the domain, which looks significantly different.
Furthermore, we propose an 
input image synthesizing method based on CutMix for cross-domains and domain identification label in discriminator for that.
Through experiments, we show the effectiveness of the proposed method using RGB images as a pre-trained model and data from multiple domains such as RGB and thermal infrared images.
%
\section{Related Work}
\subsection{Object Detection for Each Domain}
Most object detection methods have been studied for RGB images~\cite{liu2020deep,kang2022survey,wu2020recent}.
These methods can be roughly divided into two-stage and one-stage detection methods.
R-CNN and its extended technologies~\cite{girshick2014rich,Girshick2015ICCV,ren2015faster} represent the two-stage methods.
YOLO~\cite{Redmon2016CVPR}, SSD (Single Shot Multi-box Detector)~\cite{liu2016ssd}, and their extended technologies represent the one-stage methods.
Additionally, object detection techniques that use transformer have been proposed~\cite{dosovitskiy2020image,carion2020end,zhu2020deformable,yao2021efficient,liu2021swin}, and they are expected that it will be applied to various environments.
There is research on applications in the real environment such as robots~\cite{khokhlov2020tiny,szemenyei2021fully}, drones~\cite{hossain2019deep,wu2019delving}, object detection for in-vehicle cameras~\cite{lin2018graininess,zhou2018bi}, and license plate detection~\cite{xu2018towards}.
Many datasets~\cite{lin2014microsoft,Everingham15,openimages,shao2019objects365,kim2018textual,Cordts2016Cityscapes} that can be used to train object detection are available to the public.
The night scenes on these datasets are fewer than daytime scenes.
Furthermore, visibility is poor because the pixel value of the subject in the RGB image captured at night is small.
Thus, it is difficult to achieve high detection performance using RGB images both day and night.

Highly accurate object detection with in-vehicle cameras and outdoor drones is required for both day and night.
Some methods for detecting objects using spectra information other than RGB images were proposed to detect objects with high accuracy on both day and night.
Lu et al.~\cite{zhang2019weakly} proposed object detection using RGB and infrared images in the framework of weakly supervised learning.
This method focuses on the use of multispectral information, and it detects objects using a roughly aligned RGB image and an infrared image.
Liu et al.~\cite{liu2016multispectral} and Konig et al.~\cite{konig2017fully} proposed methods for inputting RGB and thermal infrared images into a deep learning model and fusing their features.
Highly accurate object detection is possible under various lighting conditions using RGB and infrared images simultaneously.
These methods are algorithms that assume that there are numerous RGB and infrared images.
Thus, they can be used if the RGB and the infrared images can be photographed in large quantities and annotated in the same environment as the inference scene.
However, the cost of collecting data and annotating in each application environment is high in reality.

\subsection{Knowledge Transfer to Different Domain}
To reduce the cost of preparing infrared images and training high accurate object detection models, domain adaptation and transfer learning use a pre-trained model with a small number of labeled infrared and RGB images.

Akkaya et al.~\cite{akkaya2021self} proposed unsupervised domain adaptation between a model taken from numerous RGB and thermal infrared images for image classification.
Vibashan et al.~\cite{vs2022meta} used paired RGB and thermal infrared images to perform domain adaptation for object detection.
When only the recognized object is displayed as in the image classification, the domains of both the difference in the sensor and shooting scene can be applied.
However, when the background area without objects occupies a large area in object detection, it is difficult to adapt the domain of both the sensor difference and the shooting scene difference without using a pair of datasets. 
Thus, it is still a problem to adapt between different sensors and scenes for object detection.

There is a knowledge transfer method, which is by fine-tuning infrared images, using a model trained with RGB images as pre-trained model.
For fine-tuning to be effective, the feature of training data between source domain and target domain must be similar.
RGB and infrared images have extremely different spectrum to be imaged, and they look very different even if they have the same object and color.
Negative transfer~\cite{pan2009survey,weiss2016survey} occurs because of the difference in the distribution of these data.
Therefore, knowledge transfer using a small number of labeled infrared images is difficult.
In both domain adaptation and fine-tuning, the key to improving performance is transferring knowledge while making the differences between domains closer.

\section{Proposed Method}
\subsection{Overview}
In this paper, we propose high-accuracy object detection on infrared images using a large number of labeled RGB images and a small number of labeled infrared images.
RGB and infrared images receive different spectra; thus, there is a significant difference in appearance, which is a large gap between domains.
Therefore, we not only align the gaps between domains at the feature level using methods such as adversarial learning but also explicitly reduce the gaps between domains at the image level.
This improves the accuracy of domain adaptation by converting the input image to conditions that make it adapt the domain easier.

\begin{figure}[t]
\centering
\includegraphics[bb=-70 0 1593 598,width=1.2\textwidth]{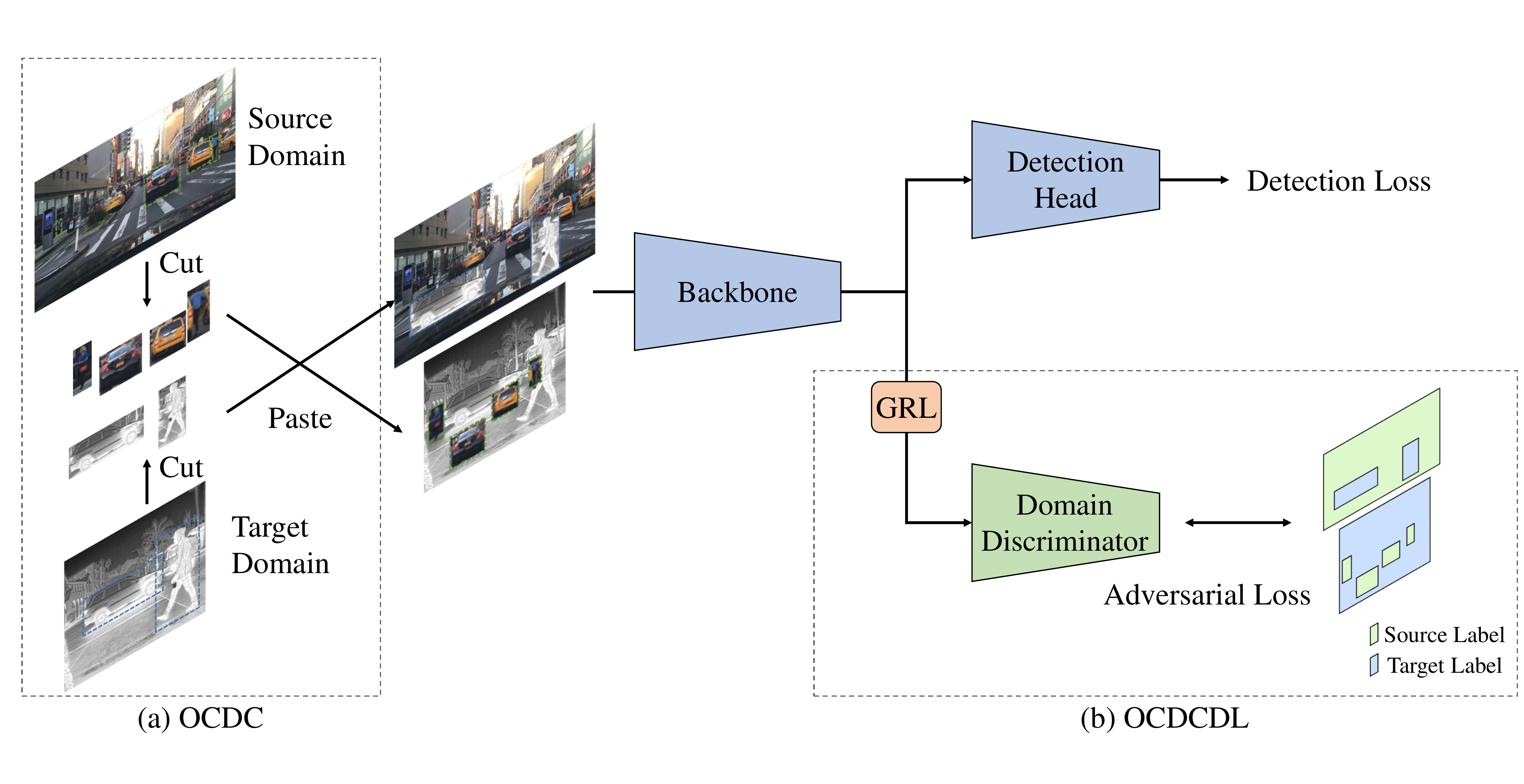}
\caption{
The framework of our proposed method;
(a) OCDC mitigates the image-level domain gap by cutting out the object area and pasting them in each other domain.
(b) We adaptively determine the domain identification label based on the area pasted by the OCDC.
The pasted object, which serve areas as new ground truth, are input into the detection network, and the detection and the adversarial losses are calculated.
}
\label{fig:overview}
\end{figure}

We propose Object aware Cross-Domain CutMix (OCDC) and OCDC-based Discriminator Label (OCDCDL) based on the domain for each location.
Figure \ref{fig:overview} shows our proposed framework.
We explain the outline of the proposed method using the model of the domain adaptation method based on adversarial learning proposed by Han-Kai et al.~\cite{hsu2020progressive} as an example.
This method trais using the loss of both object detection and adversarial learning to reduce the difference between domains.
This proposed method is simple and easy to incorporate into the type of domain adaptation that uses adversarial learning in object detection problems, which uses few labeled images.
OCDC (Fig. \ref{fig:overview} (a)) is a method for cutting out an object area and pasting a part of the image between domains to reduce the gap between the source and the target domains.
Zhou et al. showed that there is a domain generalization effect by mixing images with different domains in a batch~\cite{zhou2020domain}.
Inspired by that study, we focused on mixing object units, which is important for object detection.
When the entire image is mixed, it is trained to reduce the distance between the background domains that occupy most of the image.
In the proposed method, the distance between domains is emphasized rather than the background features, so the object detection performance is expected to improve.

OCDCDL (Fig. \ref{fig:overview} (b)) is a method for adaptively changing the domain label of the discriminator based on the pasting position of other domain images.
By converting the input image using OCDC, the domain identification label is no longer one value.
The output feature using the input image includes information from multiple domains by pasting an image of another domain on the input image using OCDC.
Thus, the conventional single identification label cannot correctly discriminate the domain.
By adaptively changing the label based on the OCDC, the discriminator makes it possible to discriminate the domain even if there is information on different domains in the image.
\subsection{Object aware Cross-Domain CutMix (OCDC)}
For domain adaptation to data with a large gap between domains such as RGB and infrared images, we propose a method of aligning domains at the image level in addition to the conventional method of aligning domains using features.
A method based on data augmentation called DomainMix augmentation~\cite{wang2020domainmix} was proposed: as a method for reducing the domain gap at the image level.
This method reduces the difference in appearance between images by simply connecting different domains without using a deep generative model.
Mixup~\cite{zhang2018mixup}, BC-learning~\cite{tokozume2018between}, CutMix~\cite{yun2019cutmix}, and CutDepth~\cite{ishii2021cutdepth} are data augmentation methods to be mixed at the image level.
These studies mention that mixing images can generate features that interpolate two images.
By mixing them at the image level, the features between the domains in the input image can be brought closer to each other.

However, the background area without objects occupies most of the image, and the objects that should align the domains are only part of the image.
Thus, in Mixup or CutMix, which uses the entire image or mixes random regions, respectively, the gap in the background domain is small, but the gap in the domain of the object to be detected is not always small.
Therefore, we propose an OCDC to cut out the object area existing in one domain and paste it onto the other domain.
The domain gap is reduced at the image level by cutting out the object area from the source and target and pasting them together.

The pasting process is performed for each training iteration.
The images of the source domain and that of the target domain included in the batch are used.
The object to be detected is cut out from those images based on the ground truth coordinates.
The object image cut out from the image of the source domain is pasted onto the image of the target domain selected at random.
On the other hand, the image cut out from the object image of the target domain is pasted onto the image of the source domain selected at random.
The ground truth labels for detection loss are updated by adding the ground truth label of the pasted image of another domain to that of the image where the image of another domain is pasted.

If the object images are pasted while the objects already exist, the originally existing objects will be hidden.
The loss of object detection becomes large because it is difficult to detect such hidden objects.
Therefore, we decide the pasting position of the image based on the overlap between the image to be pasted and the object of the image to which the object image is pasted.
If the area where the object is hidden increases more than a certain percentage after pasting the image, the pasting position is reselected when deciding the pasting position.
This prevents the original object from being hidden after pasting.

Additionally, to train the position and label of an object, the object detection model pastes an object image to a real-world location.
For example, if the domain is adapted between the images of the in-vehicle camera, which is installed at almost the same position, the coordinates before and after pasting do not change significantly.
Alternatively, the object image is pasted at the same position as the position before pasting or at a slightly shifted position.

The domain adaptation for object detection that has been used so far has insufficient consideration for object detection.
On the other hand, our method using cutting out the position of objects is a new point of view that the domain gap of the object to be recognized can be reduced.
In addition, we argue that it causes problems in adversarial learning and propose a solution.
In the following subsections, each proposed method is explained concretely.

\subsection{OCDC-based Discriminator Label (OCDCDL)}
In domain adaptation using adversarial learning such as a method~\cite{hsu2020progressive}, features calculated from the source or target domains are input into the discriminator.
In our proposed OCDC, the information of another domain is included in a part of the feature because the image of another domain is pasted on the part of the input image.
Therefore, when one value is used for the domain identification label of the discriminator as in the conventional method, the loss of the area where the image of another domain is pasted cannot be calculated correctly.
The discriminator is trained so that neither of the two domains can be discriminated.
We append the discriminator $D$ after the backbone $F$. 
The input image of the source domain and the target domain are $I_S$ and $I_T$, respectively.
$D$ outptus a domain prediction map of each pixel $D(\cdot)_{h,w}$.
The source domain identification label and target domain identification label are $d = 0$ and $d = 1$, respectively.
Eq. \ref{math:adv} is a adversarial loss $\mathcal{L}_{adv}$ and Eq. \ref{math:all} is an overall loss $\mathcal{L}$.
\begin{equation}
\label{math:adv}
\mathcal{L}_{adv}(F(I))=-\Sigma_{h,w}[d\log D(F(I))_{h,w} + (1-d)\log(1-D(F(I))_{h,w})],
\end{equation}
\begin{equation}
\label{math:all}
\min_F\max_D\mathcal{L}(I_S,I_T)=\mathcal{L}_{det}(I_S) + \mathcal{L}_{det}(I_T) + \lambda_{adv}[\mathcal{L}_{adv}(F(I_S)) + \mathcal{L}_{adv}(F(I_T))],
\end{equation}
where $\mathcal{L}_{det}(\cdot)$ is the detection loss, and $\lambda_{adv}$ is a weight that determines the loss balance.
We set $\lambda_{disc}$ is 0.1 in our experiments.

However, because the features near the boundary of the area where the object of another domain is pasted using OCDC are a mixture of the features of the two domains, it is difficult to distinguish which domain is near this boundary.
Thus, the loss near the boundary of the trained discriminator is smaller than that in other regions.
However, information from different domains is rarely mixed at a position far from the boundary of the pasted object.
Therefore, the loss of discriminator in the feature is large in that area.

Furthermore, we adaptively determine the domain identification label based on the position and domain of the image pasted by OCDC.
After the pasting process, the RGB domain label is replaced with the domain identification label corresponding to the infrared image region, and the infrared domain label is replaced with the domain identification label corresponding to the RGB image region.
%
%
\section{Experiments}
We evaluate the effectiveness of the proposed methods using RGB images~\cite{kim2018textual,dollar2009pedestrian} and thermal infrared images~\cite{fa2018flir,hwang2015multispectral} with a large domain gap.
In these experiments, there are differences in both the spectrum and the captured scene.
We compare the performance with a large amount of RGB images and a small amount of thermal infrared images.
All datasets are labeled.
Additionally, to verify the generalization performance of the proposed method, we evaluate the performance using real images ~\cite{Cordts2016Cityscapes} and simulation images ~\cite{johnson2016driving}.

\subsection{Comparison Methods}
We explain the comparison method used for each experimental setting.
In our experiment, images of the target domain are used for evaluation.
Source-only and target-only labels in the tables of experimental results show the evaluation results when the model is trained using only the image of the source or target domains, respectively.
The fine-tuning label shows the results using a pre-trained model and fine-tuning using target domain data.
The target samples label shows the number of target domain data using fine-tuning.
The Domain-Adversarial Training of Neural Networks (DANN)~\cite{ganin2016domain}, one of the adversarial learning methods, is used as the baseline of the adversarial learning method.
DANN label shows the detection results using domain adaptation with DANN.
We use Faster R-CNN~\cite{ren2015faster} as the detection network and VGG16~\cite{simonyan2014very} for the backbone.
In domain adaptation, the model parameters are pre-trained in the source domain.
The height size is 600 of the image resolution, but if the maximum width size is more than 1,000, we set it to 1,000 while maintaining the aspect ratio.
Ours label shows the result using the proposed method, which uses the OCDC and OCDCDL.
The optimizer is SGD; the learning rate, the weight decay, and the momentum are 0.001, 0.0005, and 0.9, respectively.
The batch size is one.
The evaluation metrics are the average precision at an intersection over union (IoU), which threshold is 0.5.
The front of the arrow indicates the source domain, and the tip of the arrow indicates the target domain.

\setlength{\tabcolsep}{4pt}
\begin{table}[t]
\begin{center}
\caption{
Results on BDD100k $\rightarrow$ FLIR
}
\label{table:resultflir}
\scalebox{0.85}{ %
\small %
\begin{tabular}{cccccc}
\hline\noalign{\smallskip}
Method & Target Samples & Person & Bicycle & Car & mAP\\
\noalign{\smallskip}
\hline
\hline
\noalign{\smallskip}
Source-only & --- & 39.9 & 24.9 & 68.2 & 44.4 \\
\hline
& Full & 74.2 & 57.9 & 84.1 & 72.1 \\
& 1/2 & 71.5 & 56.0 & 82.3 & 69.9 \\
& 1/4 & 66.7 & 48.7 & 78.5 & 64.6 \\
Target-only & 1/8 & 61.4 & 41.4 & 75.4 & 59.4 \\
& 1/16 & 57.0 & 42.5 & 71.8 & 57.1 \\
& 1/32 & 51.3 & 34.9 & 67.3 & 51.2 \\
& 1/64 & 44.4 & 32.0 & 63.6 & 46.7 \\
\hline
& Full & 75.0 & 60.5 & 86.3 & 73.9 \\
& 1/2 & 74.8 & 58.3 & 86.1 & 73.1 \\
& 1/4 & 72.4 & 53.1 & 85.4 & 70.1 \\
fine-tuning & 1/8 & 69.1 & 47.8 & 82.9 & 66.6 \\
& 1/16 & 64.6 & 45.4 & 79.5 & 63.2 \\
& 1/32 & 65.9 & 46.7 & \textbf{83.3} & 65.3 \\
& 1/64 & 64.5 & 41.2 & 82.0 & 62.6 \\
\hline
& Full & \textbf{78.1} & \textbf{63.8} & \textbf{87.0} & \textbf{76.3} \\
& 1/2 & 77.6 & \textbf{63.1} & \textbf{87.2} & 76.0 \\
& 1/4 & 75.2 & 56.5 & 86.2 & 72.6 \\
DANN & 1/8 & 72.4 & 58.8 & 84.5 & 71.9 \\
& 1/16 & 70.6 & 55.8 & 83.8 & 70.1 \\
& 1/32 & 69.4 & \textbf{53.8} & 82.3 & 68.5 \\
& 1/64 & 67.7 & \textbf{51.8} & 81.9 & 67.1 \\
\hline
& Full & 77.8 & 63.5 & 86.9 & 76.1\\
& 1/2 & \textbf{78.3} & 62.6 & \textbf{87.2} & \textbf{76.1} \\
& 1/4 & \textbf{76.9} & \textbf{59.9} & \textbf{86.9} & \textbf{74.5} \\
Ours & 1/8 & \textbf{75.4} & \textbf{60.9} & \textbf{85.7} & \textbf{74.0} \\
& 1/16 & \textbf{72.2} & \textbf{57.9} & \textbf{84.5} & \textbf{71.5} \\
& 1/32 & \textbf{71.1} & \textbf{53.8} & 82.0 & \textbf{69.3} \\
& 1/64 & \textbf{68.5} & 51.6 & \textbf{82.3} & \textbf{67.5} \\
\hline
\end{tabular}
}
\end{center}
\end{table}
\setlength{\tabcolsep}{1.4pt}
\textbf{BDD100k $\rightarrow$ FLIR:}
The BDD100k dataset~\cite{kim2018textual} is collected based on six types of weather conditions, six different scenes, and three categories of time of data; the number of images is 100,000.
This dataset is annotated in ten categories.
FLIR ADAS dataset~\cite{fa2018flir} is an image captured by a FLIR Tau2 camera, and the number of images is 10,228.
Only thermal infrared images from this dataset are used.
In our experiment, the training data includes 36,728 images labeled as daytime from the BDD100k dataset as the source domain data and 8,862 thermal infrared images used as training splits from the FLIR ADAS dataset. The categories person, bicycle, and car, which are common categories.

Table \ref{table:resultflir} shows the evaluation results.
The detection accuracy of source-only is the lowest because this does not use knowledge of the target domain.
In target-only, mAP is 72.1 \% using all target data named Full.
However, performances considerably deteriorates when the amount of data decreased.
In fine-tuning, performances are higher than the performances of target-only because fine-tuning models had a knowledge that performance improves somewhat even in different domains.
The performances of DANN are higher than fine-tuning because of the effects of domain adaptation.
There is no significant performance degradation due to negative transfer, but the effect of domain adaptation can be confirmed.
The performances of the proposed method tend to improve overall, but in particular, the performance of the person label outperformed DANN.

When the target samples are Full, the number of each object is large, so even if the object areas are small, there are enough numbers to improve the performance by domain adaptation.
However, in the case of Full, there is only a difference of 0.3 points, so the conventional domain adaptation does not have a difference drastically. 
In particular, in this experimental result, it should be noted that the target samples, which are our targets, are smaller than 1/2, rather than the detection result of Full.
Under these conditions, the our method has higher performance than the conventional method in almost all cases.
We confirmed about 4 point improvements over DANN even if there is a few data.
For example, if we consider person label, this is because the percentage of people in the dataset was high, and the overall percentage of pasting person's images onto another domain based on the input image was high.
\setlength{\tabcolsep}{4pt}
\begin{table}[t]
\begin{center}
\caption{
Results on Caltech $\rightarrow$ KAIST (Person) and SIM10K $\rightarrow$ Cityscapes (Car)
}
\label{table:resultkaist}
\scalebox{0.85}{ %
\small %
\begin{tabular}{cccc}
\hline\noalign{\smallskip}
Method & Target Samples & 
\begin{tabular}{c}
Person \\ (KAIST)
\end{tabular}
&
\begin{tabular}{c}
Car \\ (Cityscapes)
\end{tabular} \\
\noalign{\smallskip}
\hline
\hline
\noalign{\smallskip}
Source-only & --- & 2.8 & 43.1 \\
\hline
& Full & 67.0 & 60.3 \\
& 1/2 & 67.6 & 58.1 \\
& 1/4 & 63.0 & 54.3 \\
Target-only & 1/8 & 58.7 & 52.0 \\
& 1/16 & 57.1 & 48.3 \\
& 1/32 & 51.3 & 45.2 \\
& 1/64 & 47.4 & 41.1 \\
\hline
& Full & 63.4 & 58.1 \\
& 1/2 & 63.4 & 58.1 \\
& 1/4 & 63.5 & 57.8 \\
fine-tuning & 1/8 & 63.2 & 56.8 \\
& 1/16 & 61.7 & 55.7 \\
& 1/32 & \textbf{59.5} & 52.5 \\
& 1/64 & \textbf{57.1} & 49.8 \\
\hline
& Full & \textbf{69.4} & 62.0 \\
& 1/2 & 71.7 & 61.0 \\
& 1/4 & 70.4 & 58.6 \\
DANN & 1/8 & 69.1 & \textbf{59.6} \\
& 1/16 & 66.9 & 55.2 \\
& 1/32 & 57.0 & 54.1 \\
& 1/64 & 54.5 & 52.3 \\
\hline
& Full & 68.4 & \textbf{63.6} \\
& 1/2 & \textbf{73.3} & \textbf{61.8} \\
& 1/4 & \textbf{72.4} & \textbf{60.0} \\
Ours & 1/8 & \textbf{69.7} & \textbf{59.6} \\
& 1/16 & \textbf{67.5} & \textbf{57.4} \\
& 1/32 & 59.1 & \textbf{56.5} \\
& 1/64 & \textbf{57.1} & \textbf{54.2} \\
\hline

\end{tabular}
}
\end{center}
\end{table}
\setlength{\tabcolsep}{1.4pt}

\textbf{Caltech $\rightarrow$ KAIST:}
The Caltech Pedestrian dataset~\cite{dollar2009pedestrian} is a dataset that contains labeled images of pedestrians captured using an in-vehicle camera. 42,782 images from this dataset are used for the images.
The KAIST Multispectral Pedestrian dataset~\cite{hwang2015multispectral} is a dataset captured using the FLIR A35 microbolometer LWIR camera and contains 95,000 images labeled for pedestrians. The thermal infrared image of this dataset is used.
In KAIST, 7,688 thermal infrared images are used for training, and 2,252 thermal infrared images are used for testing.
This is based on the procedure in the paper~\cite{Zhang_2016_CVPR}.
In this experiment, only person is used as the category.

Table \ref{table:resultkaist} shows the evaluation results.
Source-only detection accuracy is extremely low because the appearance of RGB and thermal infrared images differs significantly.
Target-only and fine-tuning detection accuracy deteriorated as the target sample decreased, as demonstrated in the FLIR experiment.
Particularly, fine-tuning detection accuracy was lower than target-only detection accuracy by pre-training in the source domain.
However, the performance degradation is suppressed when the target samples decrease.
DANN and the proposed method perform better than target-only when there are many target samples.
Performance is higher than fine-tuning when the number of samples decreased.
The proposed method showed more than two points higher performance than DANN when target samples are 1/32 and 1/64.
This experiment shows that the proposed method is more effective in single-class object detection than in multi-class object detection.
In single-class object detection, when an image of another domain is pasted on an image of another domain, the image is rarely pasted to the detected object.
Therefore, few occlusion problems due to pasting occurred in the BDD100k $\rightarrow$ FLIR experiment.
Since the proposed method has a remarkable effect when the number of targets is small, it is expected to be effective in applications that are often used, such as pedestrian detection.

%
%
%
%
\textbf{SIM10K $\rightarrow$ Cityscapes:}
The SIM10K dataset~\cite{johnson2016driving} is a composite of 10,000 images generated by the Grand Theft Auto (GTA) game engine and is annotated with cars and other similar images.
The Cityscapes dataset~\cite{Cordts2016Cityscapes} consists of real images captured in multiple urban areas and segmentation labels. We used the circumscribed rectangle of the object segmentation label as the bounding box for evaluation in the car categories.
Furthermore, we used 10,000 composite images from SIM10K as a training set.
Note that 2,975 images, which are training splits from Cityscapes, are used as training data, and 500 images, which are validation splits, are used as evaluation data.
The evaluation is performed using the common category of car.

Table \ref{table:resultkaist} shows the evaluation results.
Similar to the previous results, the source-only detection accuracy is the lowest.
If the target domain detection accuracy has a large amount of data, the target-only detection performance is high, but when the amount of data is small, the performance is significantly reduced.
Fine-tuning detection accuracy reduces performance degradation 
when the amount of data is small.
However, when the amount of data is small, 
the performance of DANN improves, so the effect of domain adaptation can be confirmed.
When the target samples were 1/8, the proposed method and DANN had the same accuracy.
Cityscapes are in-vehicle camera images, and the size of the car in the image is much larger than that of a person.
Therefore, even if the number of data is reduced a little, the vehicle area can be used for domain adaptation in the same area as the background, so the accuracy is not so decrease.
In this experiment, we confirmed that the proposed method is effective not only for domain adaptation between RGB and infrared images but also for conventional problem settings.
This experiment showed that the proposed method is a general-purpose technique that can be used in various source and target data domains.
\subsection{Ablation Study}
This section shows the comparison results under different experimental conditions. In either case, a comparison is made under the evaluation conditions of BDD $\rightarrow$ FLIR.

\textbf{Contribution of Components:}
We evaluate the effect of OCDC and OCDCDL.
Table \ref{table:resultcomponents} shows the results. 
For Person, the method using both OCDC and OCDCDL had high performance.
On the other hand, Bicycle and Car have high performance even with OCDC alone.
Our experiment do not consider labels near the boundaries of objects.
Thus, even if a person with a small area makes a mistake in the discriminator label near the object boundary, 
the effect on detection accuracy is small.
However, Bicycle and Car have a large area.
Therefore, the performance decrease if the discriminator label near the object boundary is mistaken.
However, in comparison with mAP, our proposed methods have the highest accuracy and effectiveness.

\setlength{\tabcolsep}{4pt}
\begin{table}[t]
\begin{center}
\caption{
Results on contribution of components
}
\label{table:resultcomponents}
\scalebox{0.85}{ %
\small %
\begin{tabular}{ccccccc}
\hline\noalign{\smallskip}
Target Samples & OCDC & OCDCDL & Person & Bicycle & Car & mAP\\
\noalign{\smallskip}
\hline
\hline
\noalign{\smallskip}
 & & & \textbf{78.1} & \textbf{63.8} & \textbf{87.0} & \textbf{76.3} \\
Full & \checkmark & & 77.8 & 63.2 & 86.9 & 76.0 \\
 & \checkmark & \checkmark & 77.8 & 63.5 & 86.9 & 76.1 \\
\hline
 & & & 77.6 & 63.1 & 87.2 & 76.0\\
1/2 & \checkmark & & 78.2 & \textbf{63.2} & \textbf{87.4} & \textbf{76.3} \\
 & \checkmark & \checkmark & \textbf{78.3} & 62.6 & 87.2 & 76.1 \\
\hline
 & & & 75.2 & 56.5 & 86.2 & 72.6\\
1/4 & \checkmark & & 76.7 & \textbf{61.9} & 86.8 & \textbf{75.1} \\
 & \checkmark & \checkmark & \textbf{76.9} & 59.9 & \textbf{86.9} & 74.5 \\
\hline
 & & & 72.4 & 58.8 & 84.5 & 71.9\\
1/8 & \checkmark & & 74.4 & 58.5 & 85.5 & 72.8 \\
 & \checkmark & \checkmark & \textbf{75.4} & \textbf{60.9} & \textbf{85.7} & \textbf{74.0} \\
\hline
 & & & 70.6 & 55.8 & 83.8 & 70.1\\
1/16 & \checkmark & & 72.1 & 54.9 & \textbf{84.9} & 70.6 \\
 & \checkmark & \checkmark & \textbf{72.2} & \textbf{57.9} & 84.5 & \textbf{71.5} \\
\hline
 & & & 69.4 & 53.8 & 82.3 & 68.5\\
1/32 & \checkmark & & 71.0 & \textbf{54.0} & \textbf{82.6} & 69.2 \\
 & \checkmark & \checkmark & \textbf{71.1} & 53.8 & 82.0 & \textbf{69.3} \\
\hline
 & & & 67.7 & 51.8 & 81.9 & 67.1\\
1/64 & \checkmark & & 68.1 & \textbf{52.5} & 81.8 & \textbf{67.5} \\
 & \checkmark & \checkmark & \textbf{68.5} & 51.6 & \textbf{82.3} & \textbf{67.5} \\
\hline
\end{tabular}
}
\end{center}
\end{table}
\setlength{\tabcolsep}{1.4pt}
%
%
\textbf{Region Selection Strategies:}
We compare the accuracy of whether the pasting position and scale are the same before and after pasting in OCDC.
Table \ref{table:resultstrategy} shows the experimental results.
A fixed label indicates position or scale are the same before and after pasting, and a random label indicates that the position and scale are set randomly.
The detection accuracy in many cases is higher if the same position is maintained before and after pasting.
For example, in the case of an in-vehicle camera, objects are concentrated on the lower side of the image.
The object detection model trains a set of the position and the class.
To train the relationship between the position and the class, which is unlikely to occur, does not have a positive effect on the inference result.
In our experiment, by making the pasting position and size the same, the detection model is able to train the positions and scales that are likely to occur during inference.

%
%
\setlength{\tabcolsep}{4pt}
\begin{table}[t]
\begin{center}
\caption{
Results on region selection strategies
}
\label{table:resultstrategy}
\scalebox{0.85}{ %
\small %
\begin{tabular}{ccccccc}
\hline\noalign{\smallskip}
Target Samples & Position & Scaling & Person & Bicycle & Car & mAP\\
\noalign{\smallskip}
\hline
\hline
\noalign{\smallskip}
\multirow{4}{*}{Full} & Fixed & Fixed & 77.8 & \textbf{63.5} & 86.9 & \textbf{76.1} \\
 & Fixed & Random & \textbf{77.9} & 62.8 & \textbf{87.0} & 75.9 \\
 & Random & Fixed & 76.0 & 62.5 & 86.5 & 75.0 \\
 & Random & Random & 76.4 & 62.2 & 86.6 & 75.0 \\
\hline
\multirow{4}{*}{1/2} & Fixed & Fixed & \textbf{78.3} & 62.6 & 87.2 & 76.1 \\
 & Fixed & Random & \textbf{78.3} & \textbf{63.4} & \textbf{87.4} & \textbf{76.4} \\
 & Random & Fixed & 77.2 & 62.5 & 87.0 & 75.5 \\
 & Random & Random & 77.0 & 61.2 & 87.0 & 75.1 \\
\hline
\multirow{4}{*}{1/4} & Fixed & Fixed & 76.9 & 59.9 & \textbf{86.9} & 74.5 \\
 & Fixed & Random & \textbf{77.3} & \textbf{61.4} & \textbf{86.9} & \textbf{75.2} \\
 & Random & Fixed & 75.9 & 59.8 & 86.3 & 74.0 \\
 & Random & Random & 76.1 & 59.4 & 86.3 & 73.9 \\
\hline
\multirow{4}{*}{1/8} & Fixed & Fixed & \textbf{75.4} & \textbf{60.9} & \textbf{85.7} & \textbf{74.0} \\
 & Fixed & Random & 74.6 & 58.1 & 85.5 & 72.7 \\
 & Random & Fixed & 73.4 & 58.6 & 85.2 & 72.4 \\
 & Random & Random & 73.8 & 58.0 & 85.1 & 72.3 \\
\hline
\multirow{4}{*}{1/16} & Fixed & Fixed & \textbf{72.2} & \textbf{57.9} & \textbf{84.5} & \textbf{71.5} \\
 & Fixed & Random & 72.1 & 56.4 & 84.4 & 71.0 \\
 & Random & Fixed & 71.3 & 53.9 & 84.1 & 69.8 \\
 & Random & Random & 70.7 & 56.3 & 84.0 & 70.4 \\
\hline
\multirow{4}{*}{1/32} & Fixed & Fixed & \textbf{71.1} & \textbf{53.8} & 82.0 & \textbf{69.3} \\
 & Fixed & Random & 70.2 & 52.0 & \textbf{82.9} & 68.4 \\
 & Random & Fixed & 70.5 & 53.4 & 82.3 & 68.7 \\
 & Random & Random & 69.2 & 51.9 & 82.4 & 67.8 \\
\hline
\multirow{4}{*}{1/64} & Fixed & Fixed & \textbf{68.5} & \textbf{51.6} & \textbf{82.3} & \textbf{67.5} \\
 & Fixed & Random & 68.4 & 51.1 & 81.9 & 67.1 \\
 & Random & Fixed & 67.7 & 46.9 & 81.8 & 65.5 \\
 & Random & Random & 66.8 & 51.1 & 82.1 & 66.6 \\
\hline
\end{tabular}
}
\end{center}
\end{table}
\subsection{Qualitative Results}
\begin{figure}[htbp]
\centering
\scalebox{0.85}{ %
\small %
\begin{tabular}{ccccc}
\begin{minipage}{9mm}
\centering
\small %
{\scriptsize
Target\\
\scriptsize
Samples
}
\end{minipage} 
&&&&
\\
\begin{minipage}{9mm}
\centering
1/16 
\end{minipage} &
\begin{minipage}{30mm}
\centering
\includegraphics[bb=0 0 640 512,width=30mm]{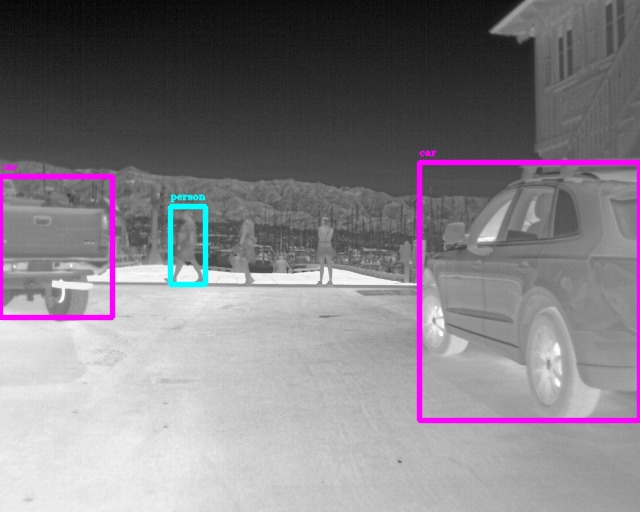}
\end{minipage} &
\begin{minipage}{30mm}
\centering
\includegraphics[bb=0 0 640 512,width=30mm]{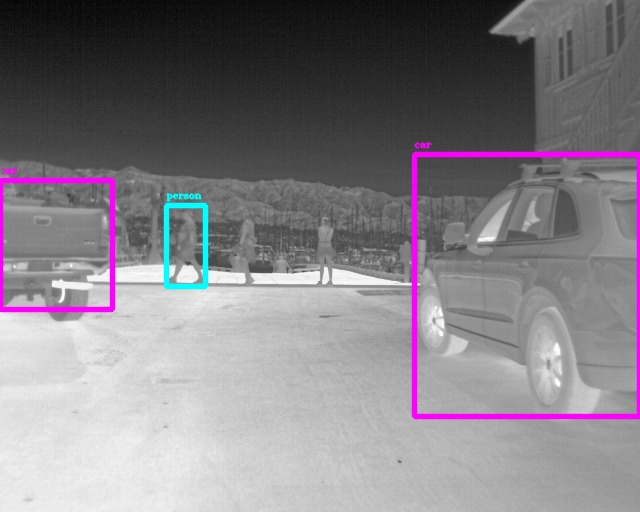}
\end{minipage} &
\begin{minipage}{30mm}
\centering
\includegraphics[bb=0 0 640 512,width=30mm]{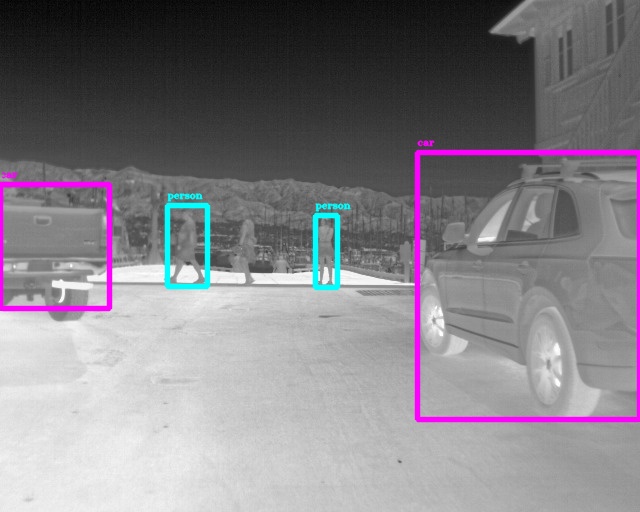}
\end{minipage} &
\begin{minipage}{30mm}
\centering
\includegraphics[bb=0 0 640 512,width=30mm]{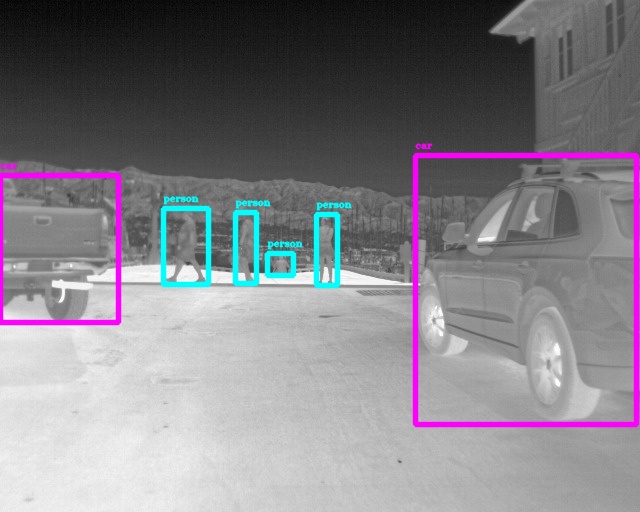}
\end{minipage} 
\\ \vspace{-3mm}
\\
\begin{minipage}{9mm}
\centering
1/64
\end{minipage} &
\begin{minipage}{30mm}
\centering
\includegraphics[bb=0 0 640 512,width=30mm]{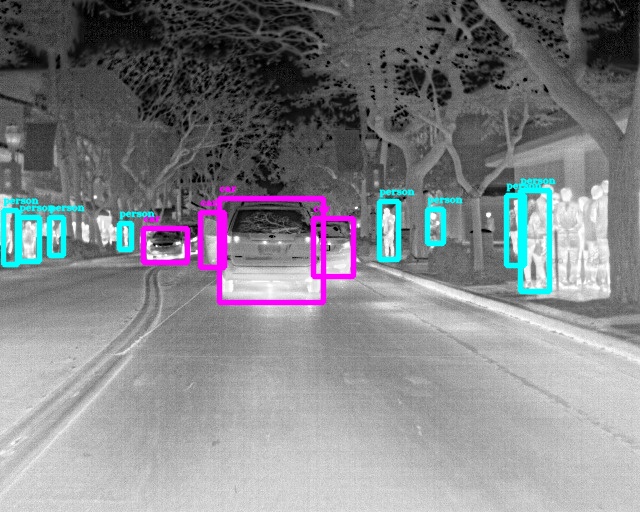}
\end{minipage} &
\begin{minipage}{30mm}
\centering
\includegraphics[bb=0 0 640 512,width=30mm]{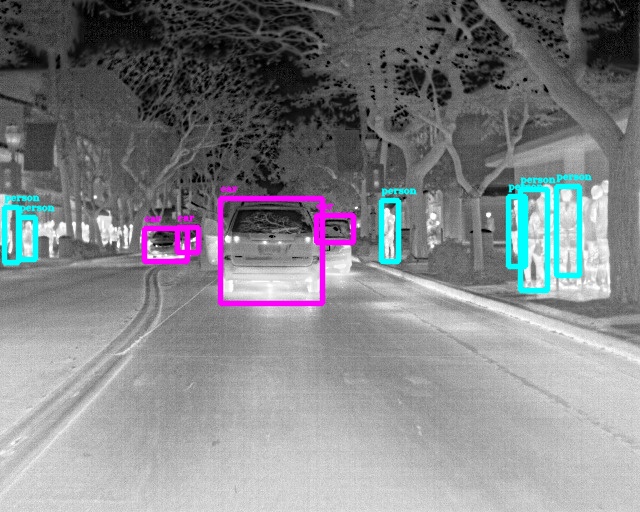}
\end{minipage} &
\begin{minipage}{30mm}
\centering
\includegraphics[bb=0 0 640 512,width=30mm]{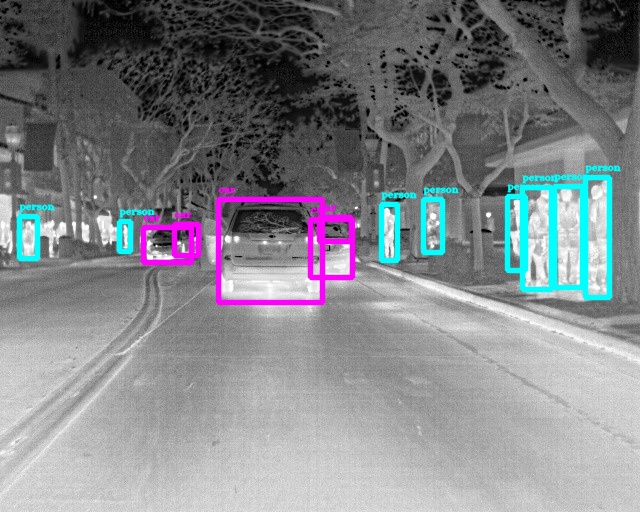}
\end{minipage} &
\begin{minipage}{30mm}
\centering
\includegraphics[bb=0 0 640 512,width=30mm]{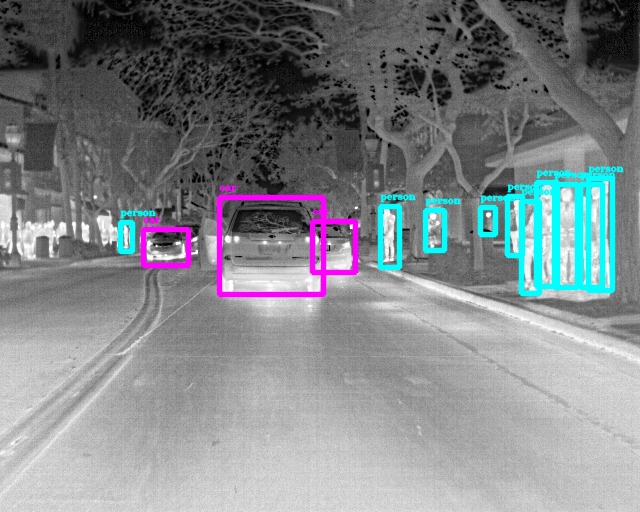}
\end{minipage} 
\\ \vspace{-3mm}
\\
&
\multicolumn{4}{c}{FLIR} \\
\begin{minipage}{9mm}
\centering
1/16
\end{minipage} &
\begin{minipage}{30mm}
\centering
\includegraphics[bb=0 0 640 512,width=30mm]{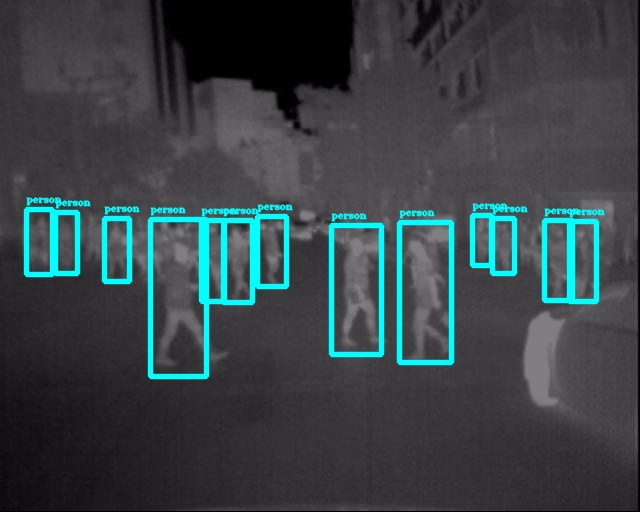}
\end{minipage} &
\begin{minipage}{30mm}
\centering
\includegraphics[bb=0 0 640 512,width=30mm]{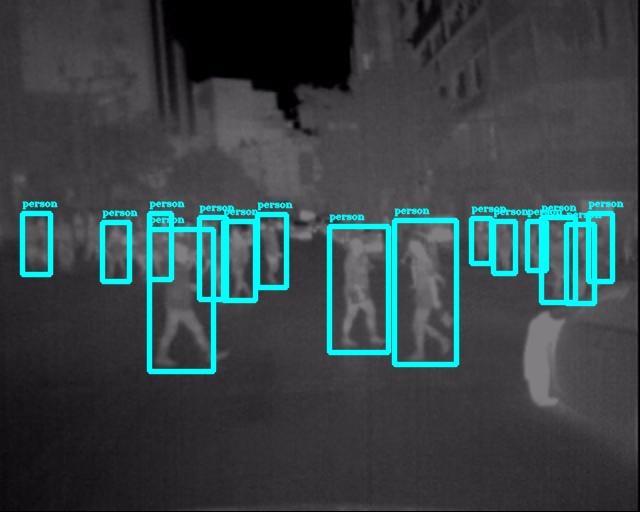}
\end{minipage} &
\begin{minipage}{30mm}
\centering
\includegraphics[bb=0 0 640 512,width=30mm]{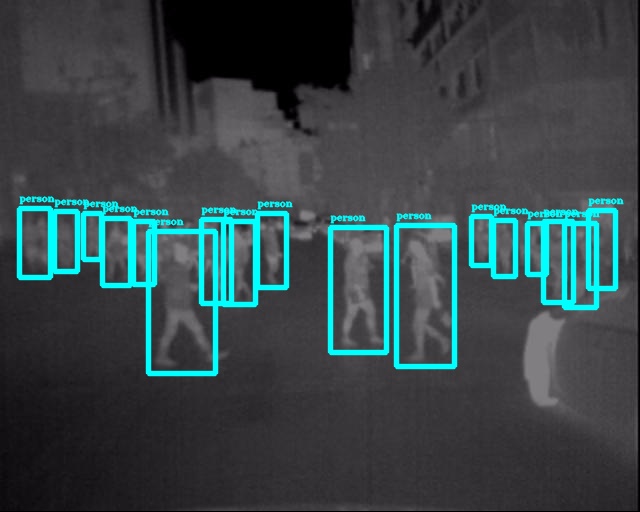}
\end{minipage} &
\begin{minipage}{30mm}
\centering
\includegraphics[bb=0 0 640 512,width=30mm]{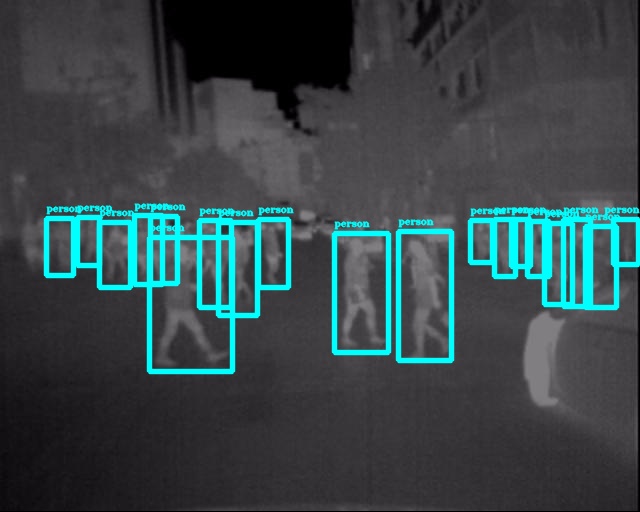}
\end{minipage} 
\\ \vspace{-3mm}
\\
\begin{minipage}{9mm}
\centering
1/64
\end{minipage} &
\begin{minipage}{30mm}
\centering
\includegraphics[bb=0 0 640 512,width=30mm]{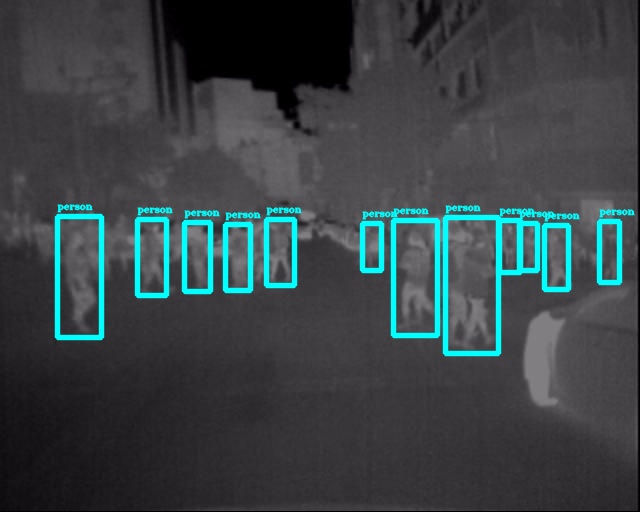}
\end{minipage} &
\begin{minipage}{30mm}
\centering
\includegraphics[bb=0 0 640 512,width=30mm]{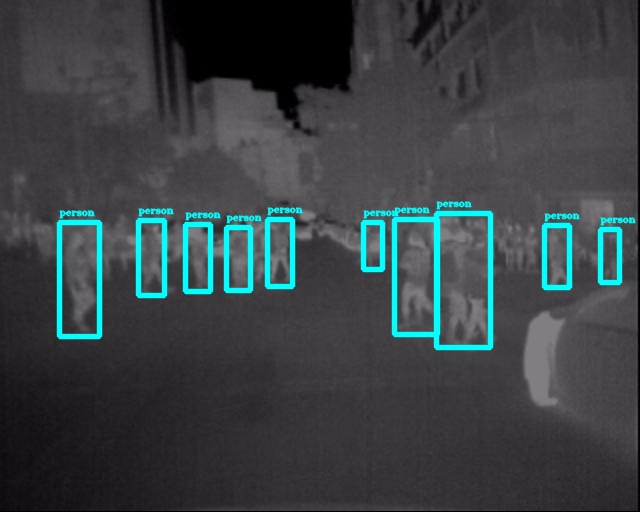}
\end{minipage} &
\begin{minipage}{30mm}
\centering
\includegraphics[bb=0 0 640 512,width=30mm]{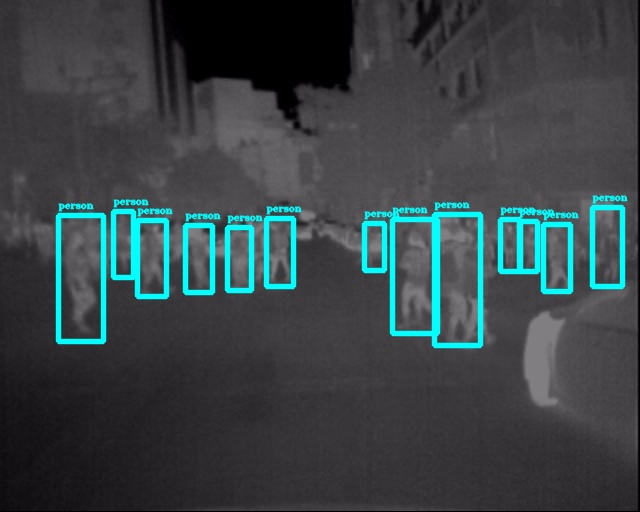}
\end{minipage} &
\begin{minipage}{30mm}
\centering
\includegraphics[bb=0 0 640 512,width=30mm]{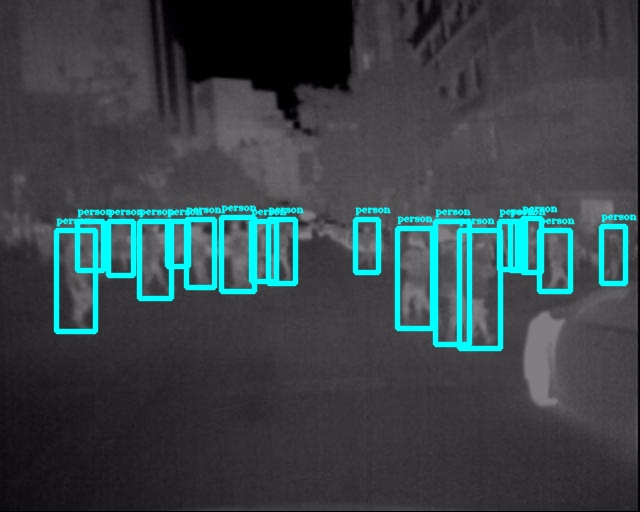}
\end{minipage} 
\\ \vspace{-3mm}
\\
&
\multicolumn{4}{c}{KAIST} \\
\begin{minipage}{9mm}
\centering
1/16
\end{minipage} &
\begin{minipage}{30mm}
\centering
\includegraphics[bb=0 0 2048 1024,width=30mm]{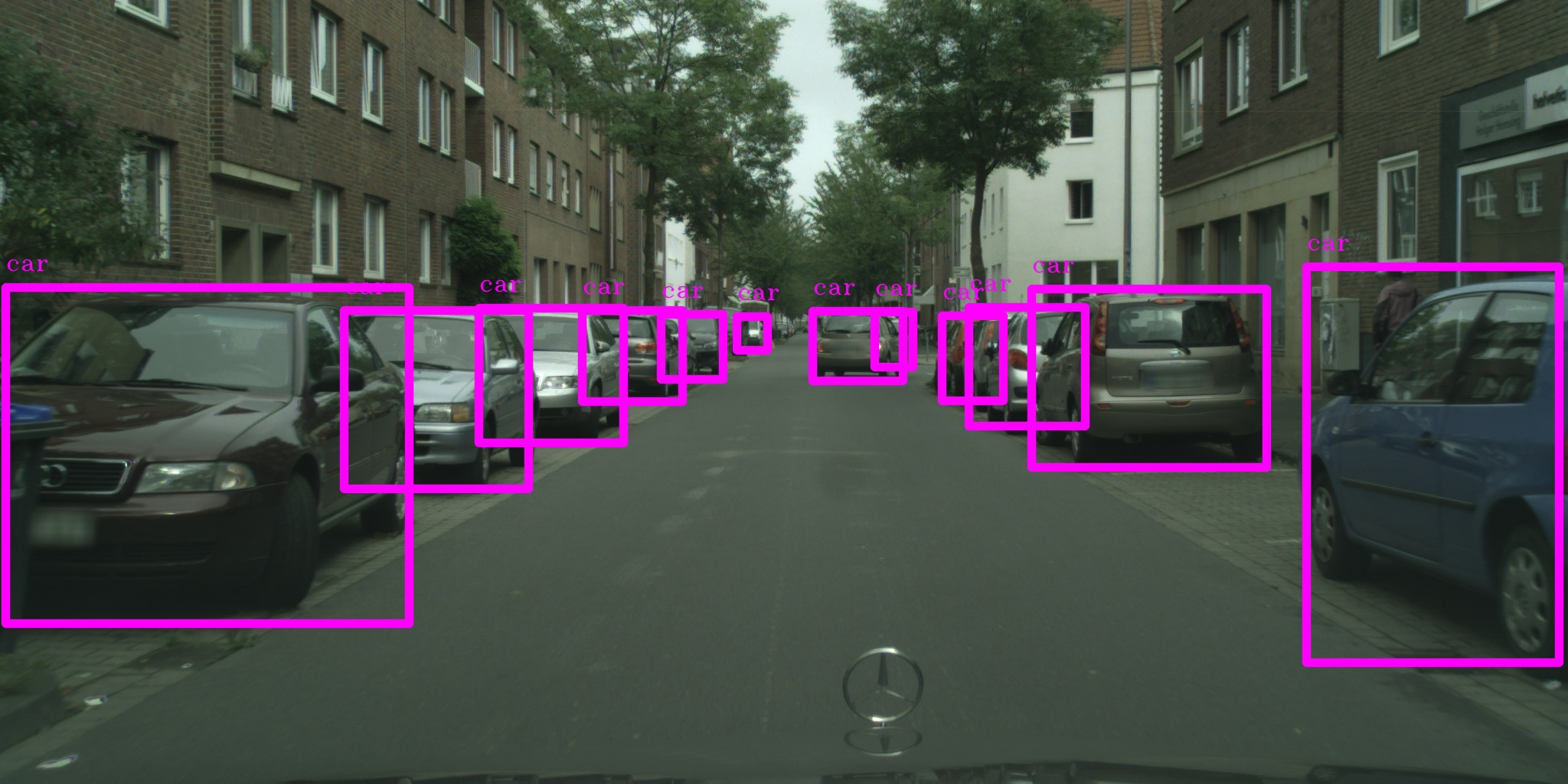}
\end{minipage} &
\begin{minipage}{30mm}
\centering
\includegraphics[bb=0 0 2048 1024,width=30mm]{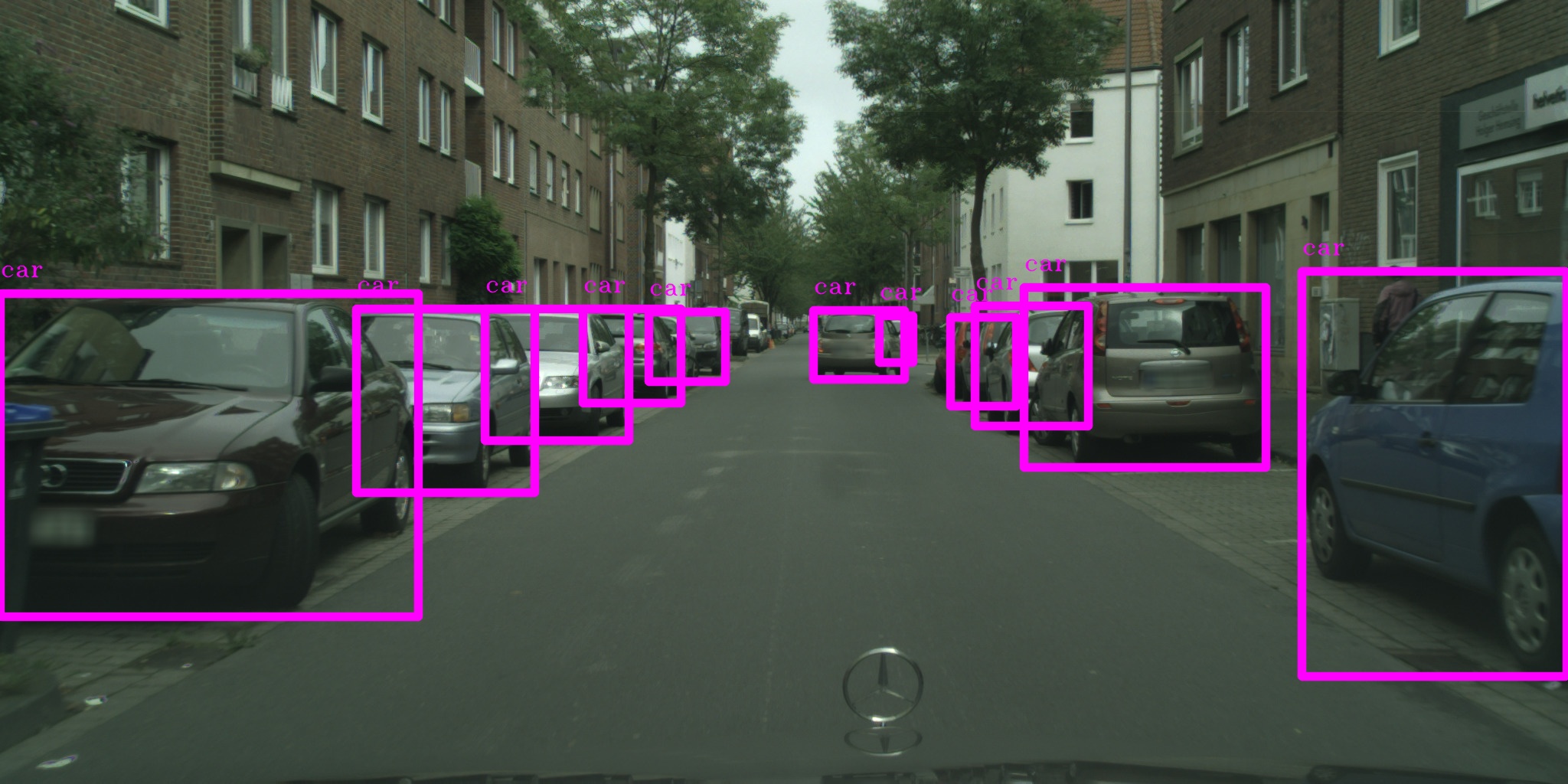}
\end{minipage} &
\begin{minipage}{30mm}
\centering
\includegraphics[bb=0 0 2048 1024,width=30mm]{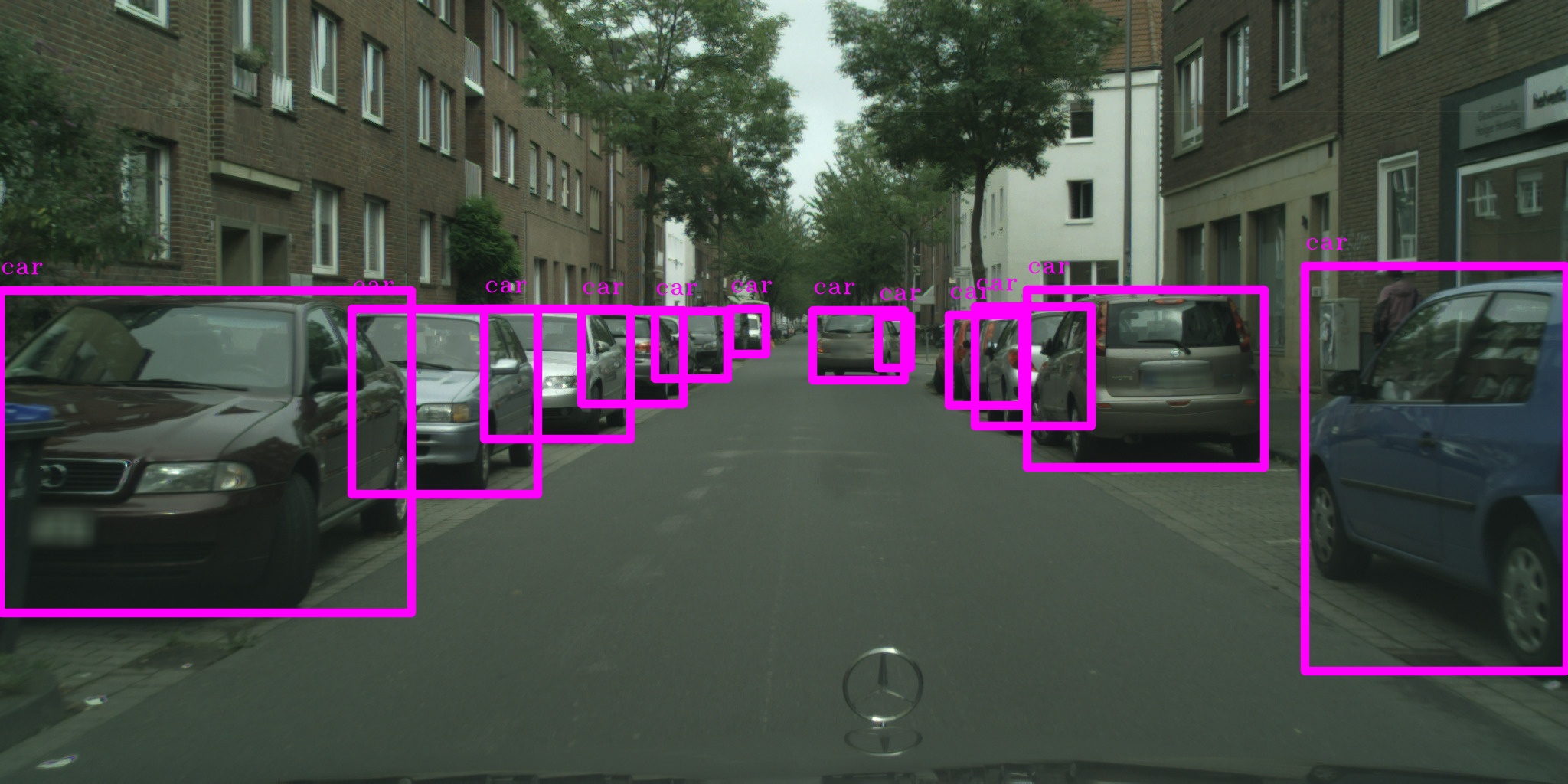}
\end{minipage} &
\begin{minipage}{30mm}
\centering
\includegraphics[bb=0 0 2048 1024,width=30mm]{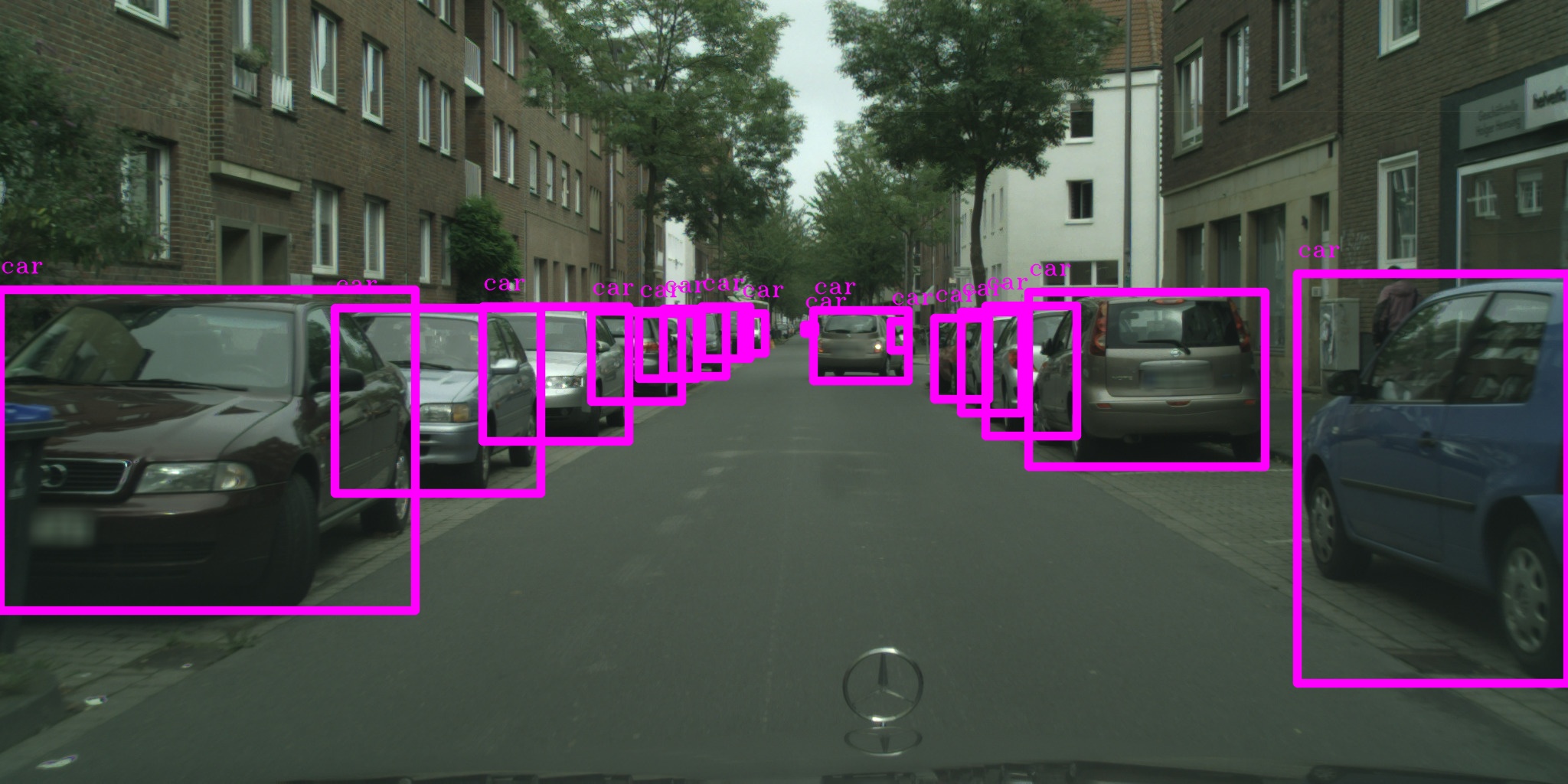}
\end{minipage} 
\\ \vspace{-3mm}
\\
\begin{minipage}{9mm}
\centering
1/64
\end{minipage} &
\begin{minipage}{30mm}
\centering
\includegraphics[bb=0 0 2048 1024,width=30mm]{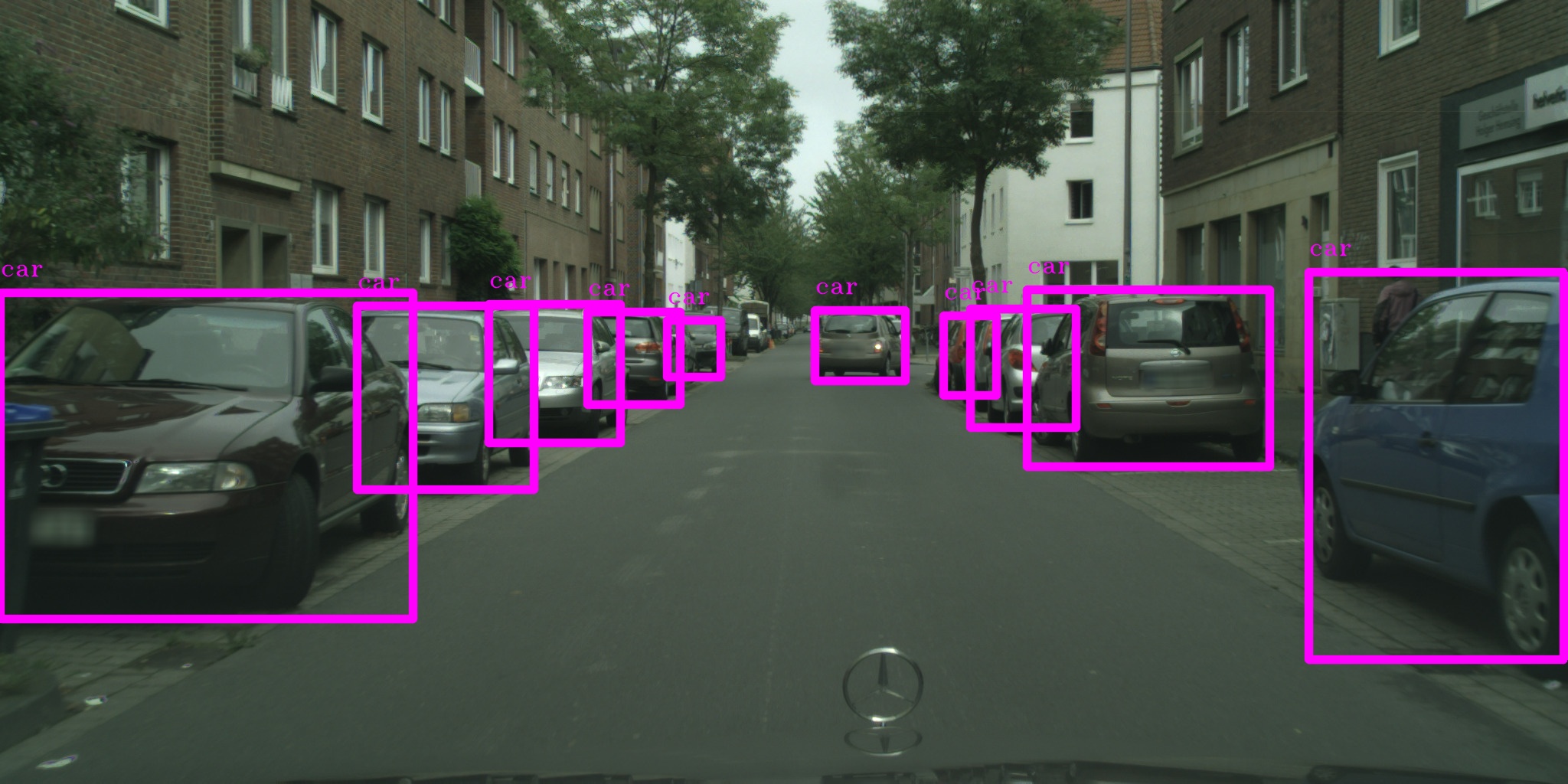}
\end{minipage} &
\begin{minipage}{30mm}
\centering
\includegraphics[bb=0 0 2048 1024,width=30mm]{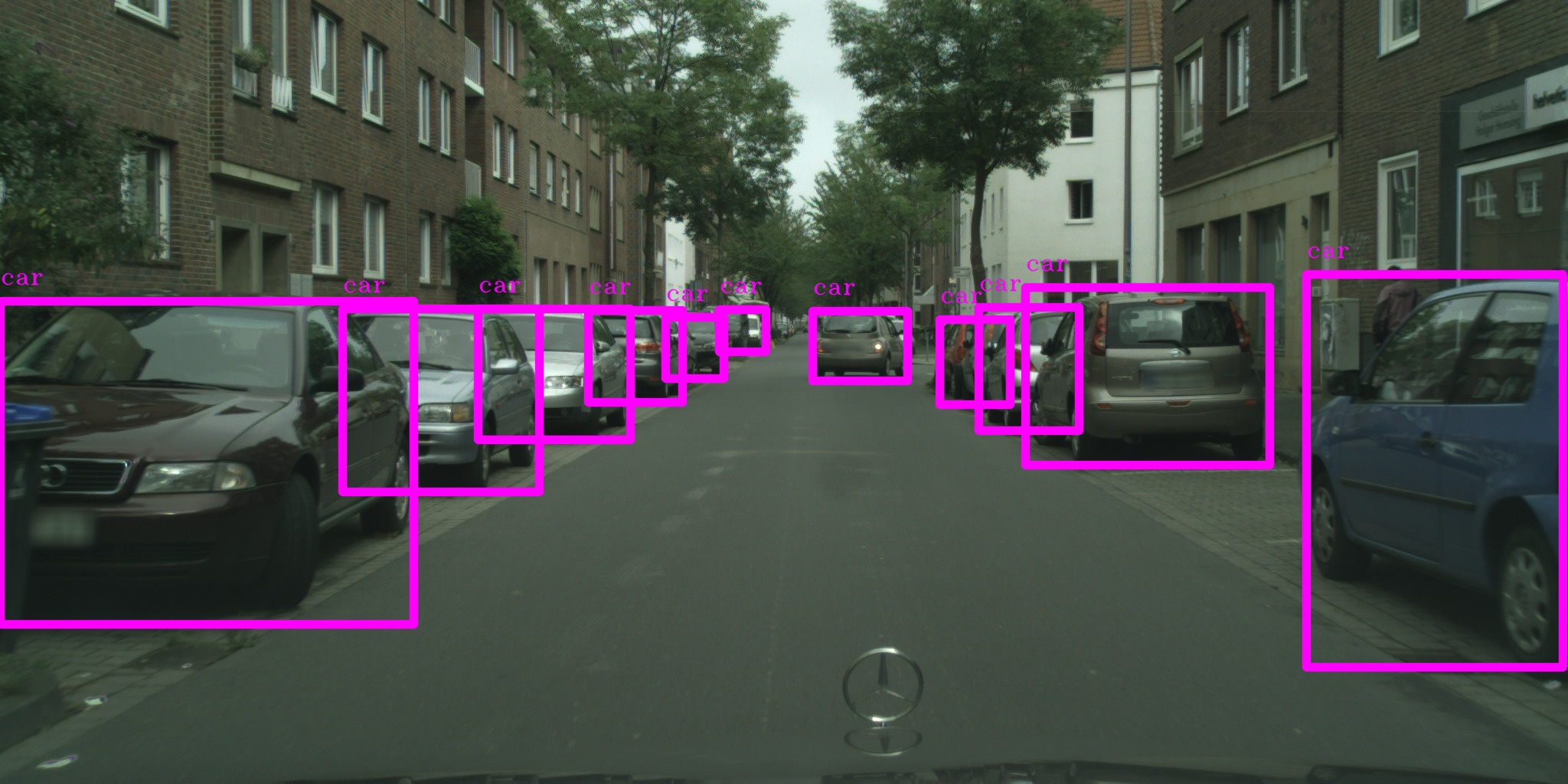}
\end{minipage} &
\begin{minipage}{30mm}
\centering
\includegraphics[bb=0 0 2048 1024,width=30mm]{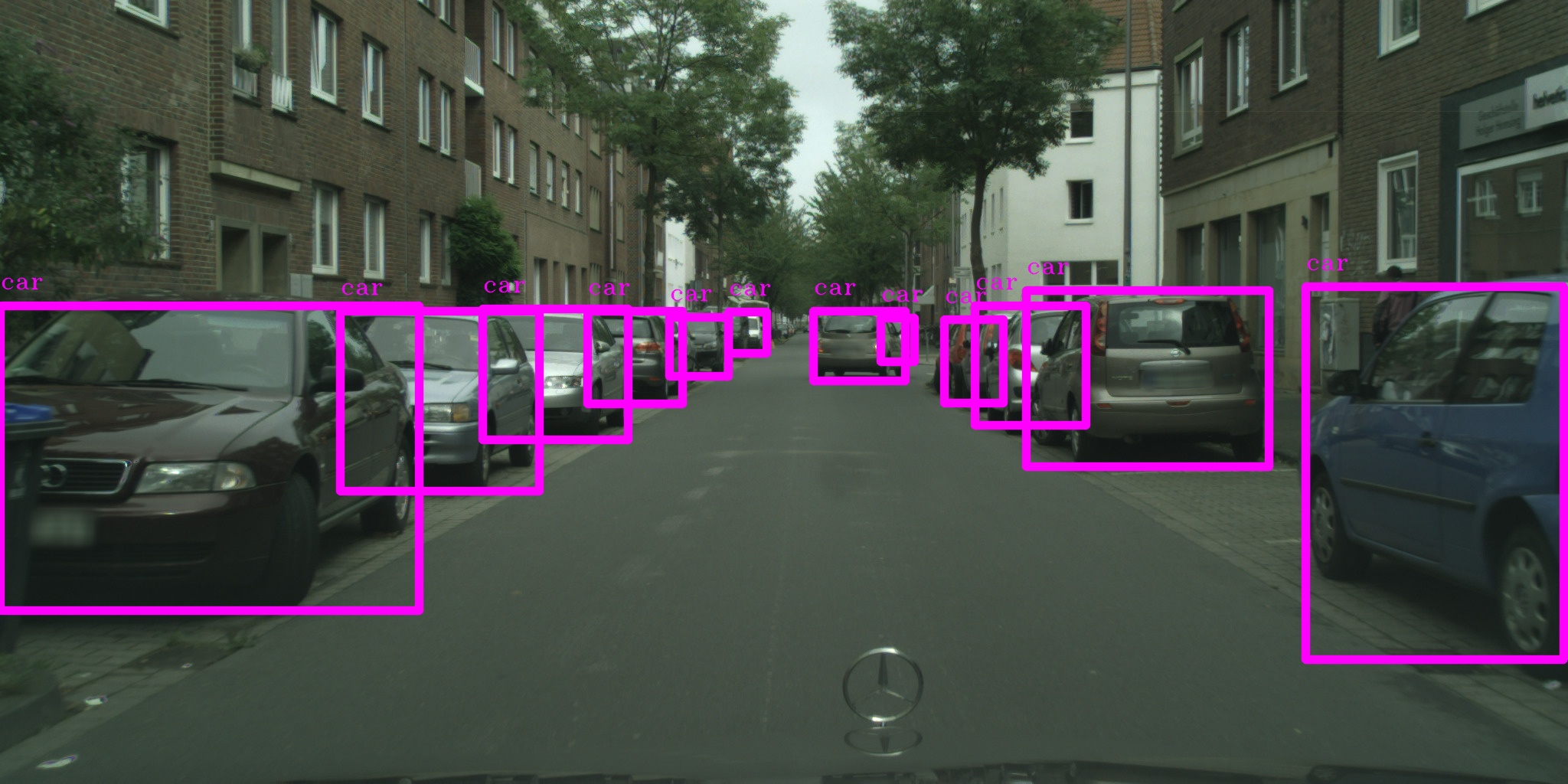}
\end{minipage} &
\begin{minipage}{30mm}
\centering
\includegraphics[bb=0 0 2048 1024,width=30mm]{figures/munster_000016_000019_leftImg8bit_blurred_gt.jpg}
\end{minipage} \\
&
\multicolumn{4}{c}{Cityscapes} \\
&
\begin{minipage}[b]{30mm}
\centering
\subcaption{fine-tuning}
\end{minipage} &
\begin{minipage}[b]{30mm}
\centering
\subcaption{DANN}
\end{minipage} &
\begin{minipage}[b]{30mm}
\centering
\subcaption{Ours}
\end{minipage} &
\begin{minipage}[b]{30mm}
\centering
\subcaption{Ground Truth}
\end{minipage} \\
\end{tabular}
}
\caption{
There are the detection results using (a) fine-tuning, (b) DANN, (c) our method, and (d) ground truth, respectively.
The first and second rows are the detection results of FLIR images using BDD100k $\rightarrow$ FLIR adaptation,
The third and fourth rows are the detection results of KAIST images using Caltech $\rightarrow$ KAIST adaptation,
The fifth and sixth rows are the detection results of Cityscapes using SIM10K $\rightarrow$ Cityscapes. 
The odd rows are 1/16 target samples and the even rows are 1/64, respectively.
The bounding box colored cyan indicates person, and the bounding box colored magenta indicates car, respectively.
}
\label{fig:result}
\end{figure}

Figure \ref{fig:result} is object detection results of FLIR, KAIST, and Cityscapes after domain adaptation performed in subsection 4.1.
At the top of each dataset is the result when target samples is 1/16, and at the bottom of each dataset is the result when it is 1/64.
The comparison methods are (a) fine-tuning, (b) DANN, and (c) Ours, respectively, and (d) Ground Truth.
The car detection result is shown in magenta, and the person detection result is shown in cyan.

In the FLIR results, there is no difference in the car detection results, but there is a difference in the person detection results.
In fine-tuning and DANN, even people with similar reflection intensities in thermal infrared images are not detected.
On the other hand, the proposed method detects objects with similar reflection intensities, even if they are people far away.
The proposed method adapted the domain to information about the reflection intensity of persons, which is a small area in the image. 
On the other hand, the conventional methods did not fully adapt the domain, so some objects could not be detected.

In the KAIST results, the conventional methods did not detect some small persons.
In domain adaptation, it is difficult to adapt information on small objects because even if the information in a small object is ignored, the loss of the detection model decreases.
However, the proposed method makes it easier to detect even small objects by explicitly giving information from other domains to the input image.

In the Cityscapes results, in both 1/16 and 1/64, the farthest car on the left side was not detected by conventional methods, but the proposed method could detect the car.
This is because it is difficult to adapt the domain of a small object, which is the same reason as in the case of KAIST.
In this experiment, we clarified the importance of explicitly giving information of other domains to the input image in domain adaptation of small objects as in the proposed method.

%
%
%
%
%
\section{Conclusion}
We proposed few-shot supervised domain adaptation for object detection in cases with large domain gaps, such as RGB and thermal infrared images.
Although the number of infrared images is smaller than that of RGB images, the performance is improved compared with the conventional domain identification via OCDC method for reducing the gap between domains for the input image we proposed and the corresponding change in the domain identification label of the discriminator (OCDCDL). 
Furthermore, it was confirmed that the proposed method is effective by a comparative experiments in which the number of data was changed, and the versatility of the method was shown by a comparative experiments using various data.
\newpage

\bibliographystyle{splncs}
\bibliography{egbib}

\end{document}


\pagestyle{headings}
\mainmatter

\def\ACCV22SubNumber{296} 

\title{Supplementary Meterial:\\Few-shot Adaptive Object Detection\\
with Cross-Domain CutMix}
\titlerunning{ACCV-22 submission ID \ACCV22SubNumber}
\authorrunning{ACCV-22 submission ID \ACCV22SubNumber}

\author{Anonymous ACCV 2022 submission}
\institute{Paper ID \ACCV22SubNumber}

\maketitle

\appendix
\renewcommand{\thefigure}{A\arabic{figure}}
\renewcommand{\thetable}{\Alph{section}.\arabic{table}}
\renewcommand{\thealgocf}{\Alph{section}.\arabic{algocf}}

\section{Pseudocode of Method}
We show the pseudocode of the procedure for applying OCDC and OCDCDL from the target domain to the source domain in Algorithm \ref{alg:ocdc}.
While we show the procedure in the direction from the target domain to the source domain, a similar procedure is used from the source domain to the target domain, with the labels $S$ and $T$ swapped.

\begin{algorithm}
\caption{The procedure of our proposed method}
\label{alg:ocdc}
\KwIn{Images of source and target domain $I_{S}$, $I_{T}$, bounding boxes of source and target domain $B_{S}$, $B_{T}$, domain identification labels $\mathcal{D}_{S}$, $\mathcal{D}_{T}$ in a batch.}
\KwOut{Pasted images of source and target domain $\hat{I_{S}}$, $\hat{I_{T}}$, added bounding boxes of source and target domain $\hat{B_{S}}$, $\hat{B_{T}}$, replaced domain identification labels $\hat{\mathcal{D}_{S}}$, $\hat{\mathcal{D}_{T}}$ in a batch.}
\textbf{Difinition:} $(x, y, w, h)$: upper left position, width, and height of $b$, $\gamma$: overlap threshold, $(W, H)$: width and height of image.\\
\ForAll{\textup{bounding box of target domain image $b_{T} \in B_{T}$}}{
// selecting a bounding box matched size criteria\\
\If{$16 < w_{b_{T}} < W_{I_{S}}$ and $16 < h_{b_{T}} < H_{I_{S}}$}{
// setting $\hat{w_{b_{T}}}, \hat{h_{b_{T}}}$, and left upper position $\hat{x_{b_{T}}}, \hat{y_{b_{T}}}$\\
$s_{rand} \sim \mathcal{U}(0.7, 1.3)$\\
$\hat{w_{b_{T}}} \gets w_{b_{T}} \mathrel{*} s_{rand}$\\
$\hat{h_{b_{T}}} \gets h_{b_{T}} \mathrel{*} s_{rand}$\\
$\hat{x_{b_{T}}} \sim \mathcal{U}(0, W_{I_{S}} \mathrel{-} \hat{w_{b_{T}}})$\\
$\hat{y_{b_{T}}} \sim \mathcal{U}(0, H_{I_{S}} \mathrel{-} \hat{h_{b_{T}}})$\\
\ForAll{\textup{bounding box of source domain image $b_{S} \in B_{S}$}}{
// calculating intersection area between $\hat{b_{T}}$ and $b_{S}$\\
$\mathcal{R}_{inter} \gets intersection(\hat{b_{T}}, b_{S})$\\
// calculating area of $b_{S}$\\
$\mathcal{R}_{Sarea} \gets area(b_{S})$\\
$\mathcal{R}_{overlap} \gets \mathcal{R}_{inter} \mathrel{/} \mathcal{R}_{Sarea}$\\
\If{$\mathcal{R}_{overlap} > \gamma$}{
\textup{go to $\hat{w_{b_{T}}}, \hat{h_{b_{T}}}, \hat{x_{b_{T}}}, \hat{y_{b_{T}}}$ setting (line 6)}
}
}
// cropping region $\mathcal{A}_{T}$ from $I_{T}$\\
$\mathcal{A}_{T} \gets crop(I_{T}, b_{T})$\\
// resizing region $\mathcal{A}_{T}$ to $\hat{w_{b_{T}}}, \hat{h_{b_{T}}}$\\
$\hat{\mathcal{A}_{T}} \gets resize(\mathcal{A}_{T}, \hat{w_{b_{T}}}, \hat{h_{b_{T}}})$\\
// pasting $\hat{\mathcal{A}_{T}}$ on $\hat{b_{T}}$ region of $I_{S}$\\
$\hat{I_{S}} \gets paste(I_{S}, \hat{\mathcal{A}_{T}}, \hat{b_{T}})$\\
// adding $\hat{b_{T}}$ to $B_{S}$\\
$\hat{B_{S}} \gets add(B_{S}, \hat{b_{T}})$\\
// switching domain identification labels corresponding to $\hat{b_{T}}$\\
$\hat{\mathcal{D}_{S}} \gets switch(\mathcal{D}_{S}, \hat{b_{T}} \mathrel{/} 16)$\\
}
}
\end{algorithm}

\section{Additional Results}
When pasting an object in image A into image B, the overlap between the objects in image A and image B is considered.
For consideration, the overlap ratio between the pasted object and the pasting object is calculated.
If the overlap is exceeded more than threshold $\gamma$, the pasting is not performed.
Table \ref{table:resultiouthresh} shows the performance comparison result.
When $\gamma$ is made smaller, the allowance for overlapping objects become stricter, and the number of objects that cannot be pasted increases.
There is no big difference overall even if $\gamma$ is changed.
Therefore, we calculate the average of mAP for each target sample and the difference between the average of mAP and the mAP for each $\gamma$ setting in Table \ref{table:resultaverage}.
The greater the difference between each $\gamma$ in the positive direction, the more beneficial $\gamma$ is for improving accuracy in that target samples.
For example, when target samples label is full, the average of mAP is 75.7 \%, and differences from the average of mAP are 0.2 point at $\gamma = 0.1$, 0.4 point at $\gamma = 0.2$, -0.1 point at $\gamma = 0.5$, -0.6 point at $\gamma = 0.75$, respectively.
When the average difference points calculated for all target samples labels at each $\gamma$ are added, the total is 1.3 points at $\gamma = 0.25$, indicating a higher performance.
Thus, we set the optimal $\gamma$ to 0.25 in the experiments.
\begin{table}[htbp]
  \begin{center}
    \caption{
      Results on the overlap threshold $\gamma$
    }
    \label{table:resultiouthresh}
    \scalebox{0.85}
    {
      \begin{tabular}{cccccc}
        \hline\noalign{\smallskip}
        Target Samples & $\gamma$ & Person & Bicycle & Car & mAP\\
        \noalign{\smallskip}
        \hline
        \hline
        \noalign{\smallskip}
        \multirow{4}{*}{Full}& 0.1 & \textbf{77.8} & 62.8 & \textbf{87.1} & 75.9 \\
        & 0.25 & \textbf{77.8} & \textbf{63.5} & 86.9 & \textbf{76.1} \\
        & 0.5 & 77.4 & 62.6 & 86.8 & 75.6 \\
        & 0.75 & 76.8 & 61.5 & 87.0 & 75.1 \\
        \hline
        \noalign{\smallskip}
        \multirow{4}{*}{1/2} & 0.1 & 78.2 & \textbf{64.1} & \textbf{87.4} & \textbf{76.6} \\
        & 0.25 & 78.3 & 62.6 & 87.2 & 76.1 \\
        & 0.5 & 78.3 & 62.4 & 87.2 & 76.0 \\
        & 0.75 & \textbf{78.4} & 63.1 & 87.2 & 76.3 \\
        \hline
        \noalign{\smallskip}
        \multirow{4}{*}{1/4} & 0.1 & 76.7 & \textbf{61.4} & 86.8 & 75.0 \\
        & 0.25 & 76.9 & 59.9 & 86.9 & 74.5 \\
        & 0.5 & 77.4 & 61.0 & 86.9 & \textbf{75.1} \\
        & 0.75 & \textbf{77.6} & 59.4 & \textbf{87.1} & 74.7 \\
        \hline
        \noalign{\smallskip}
        \multirow{4}{*}{1/8} & 0.1 & 75.0 & 59.5 & \textbf{85.8} & 73.4 \\
        & 0.25 & \textbf{75.4} & \textbf{60.9} & 85.7 & \textbf{74.0} \\
        & 0.5 & 74.8 & 58.9 & 85.5 & 73.1 \\
        & 0.75 & 75.1 & 58.8 & 85.6 & 73.2 \\
        \hline
        \noalign{\smallskip}
        \multirow{4}{*}{1/16} & 0.1 & 72.3 & 56.2 & 84.2 & 70.9 \\
        & 0.25 & 72.2 & \textbf{57.9} & 84.5 & \textbf{71.5} \\
        & 0.5 & 72.3 & 54.3 & \textbf{84.8} & 70.5 \\
        & 0.75 & \textbf{72.4} & 55.4 & 84.5 & 70.8 \\
        \hline
        \noalign{\smallskip}
        \multirow{4}{*}{1/32}&0.1 & 70.4 & 53.7 & 82.9 & 69.0 \\
        & 0.25 & 71.1 & 53.8 & 82.0 & 69.3 \\
        & 0.5 & 70.2 & 54.6 & 83.2 & 69.3 \\
        & 0.75 & \textbf{71.3} & \textbf{55.8} & \textbf{83.5} & \textbf{70.2} \\
        \hline
        \noalign{\smallskip}
        \multirow{4}{*}{1/64}&0.1 & 68.0 & \textbf{51.9} & 81.6 & 67.2 \\
        & 0.25 & 68.5 & 51.6 & 82.3 & \textbf{67.5} \\
        & 0.5 & \textbf{68.7} & 49.7 & \textbf{82.4} & 66.9 \\
        & 0.75 & 68.0 & 50.4 & 82.0 & 66.8 \\
        \hline
      \end{tabular}
    }
  \end{center}
\end{table}

\begin{table}[htbp]
  \begin{center}
    \caption{
      Results on the average of mAP and the difference from the average of mAP
    }
    \label{table:resultaverage}
    \scalebox{0.85}
    {
      \begin{tabular}{cccc}
        \hline\noalign{\smallskip}
        Target Samples & 
        \begin{tabular}{c}
        the average \\ of mAP
        \end{tabular} & $\gamma$ &
        \begin{tabular}{c}
        the difference from \\ the average of mAP
        \end{tabular} \\
        \noalign{\smallskip}
        \hline
        \hline
        \noalign{\smallskip}
        \multirow{4}{*}{Full} & \multirow{4}{*}{75.7} & 0.1 & +0.2 \\
        & & 0.25 & \textbf{+0.4} \\
        & & 0.5 & -0.1 \\
        & & 0.75 & -0.6 \\
        \hline
        \noalign{\smallskip}
        \multirow{4}{*}{1/2} & \multirow{4}{*}{76.2} & 0.1 & \textbf{+0.3} \\
        & & 0.25 & -0.2 \\
        & & 0.5 & -0.2 \\
        & & 0.75 & \pm0.0 \\
        \hline
        \noalign{\smallskip}
        \multirow{4}{*}{1/4} & \multirow{4}{*}{74.8} & 0.1 & +0.2 \\
        & & 0.25 & -0.3 \\
        & & 0.5 & \textbf{+0.3} \\
        & & 0.75 & -0.1 \\
        \hline
        \noalign{\smallskip}
        \multirow{4}{*}{1/8} & \multirow{4}{*}{73.4} & 0.1 & \pm0.0 \\
        & & 0.25 & \textbf{+0.6} \\
        & & 0.5 & -0.3 \\
        & & 0.75 & -0.2 \\
        \hline
        \noalign{\smallskip}
        \multirow{4}{*}{1/16} & \multirow{4}{*}{70.9} & 0.1 & \pm0.0 \\
        & & 0.25 & \textbf{+0.6} \\
        & & 0.5 & -0.4 \\
        & & 0.75 & -0.1 \\
        \hline
        \noalign{\smallskip}
        \multirow{4}{*}{1/32} & \multirow{4}{*}{69.4} & 0.1 & -0.4 \\
        & & 0.25 & -0.2 \\
        & & 0.5 & -0.1 \\
        & & 0.75 & \textbf{+0.7} \\
        \hline
        \noalign{\smallskip}
        \multirow{4}{*}{1/64} & \multirow{4}{*}{67.1} & 0.1 & +0.1 \\
        & & 0.25 & \textbf{+0.4} \\
        & & 0.5 & -0.2 \\
        & & 0.75 & -0.3 \\
        \hline
      \end{tabular}
    }
  \end{center}
\end{table}